\documentclass[sigconf]{aamas}

\setcopyright{ifaamas}

\setcopyright{none}
\acmConference[ALA '22]{Proc.\@ of the Adaptive and Learning Agents Workshop (ALA 2022)}
{May 9-10, 2022}{Online, \url{https://ala2022.github.io/}}{Cruz, Hayes, Silva, Santos (eds.)}
\copyrightyear{2022}
\acmYear{2022}
\acmDOI{}
\acmPrice{}
\acmISBN{}
\acmSubmissionID{12}

\usepackage[english]{babel}
\usepackage{blindtext}
\usepackage{subcaption}
\usepackage{hyperref}
\usepackage{balance}

\title[Assessing Human Interaction With Continually Learning Prediction Agents]{Assessing Human Interaction in Virtual Reality With Continually Learning Prediction Agents Based on Reinforcement Learning Algorithms: A Pilot Study}

\author{Dylan J. A. Brenneis}
\orcid{0000-0002-6359-0832}
\affiliation{%
  \institution{DeepMind Technologies, Ltd.}
  \streetaddress{10065 Jasper Avenue, Suite 250}
  \city{Edmonton}
  \state{Alberta}
  \country{Canada}
  \postcode{T5J 3B1}
}
\email{brenneis@deepmind.com}

\author{Adam S. R. Parker}
\affiliation{%
  \institution{University of Alberta}
  \city{Edmonton}
  \state{Alberta}
  \country{Canada}
}
\email{asparker@ualberta.ca}

\author{Michael Bradley Johanson}
\affiliation{%
  \institution{DeepMind Technologies, Ltd.}
  \streetaddress{10065 Jasper Avenue, Suite 250}
  \city{Edmonton}
  \state{Alberta}
  \country{Canada}
  \postcode{T5J 3B1}
}
\email{mjohanson@deepmind.com}

\author{Andrew Butcher}
\affiliation{%
  \institution{DeepMind Technologies, Ltd.}
  \streetaddress{10065 Jasper Avenue, Suite 250}
  \city{Edmonton}
  \state{Alberta}
  \country{Canada}
  \postcode{T5J 3B1}
}
\email{abutcher@deepmind.com}

\author{Elnaz Davoodi}
\affiliation{%
  \institution{DeepMind Technologies, Ltd.}
  \streetaddress{10065 Jasper Avenue, Suite 250}
  \city{Edmonton}
  \state{Alberta}
  \country{Canada}
  \postcode{T5J 3B1}
}
\email{elnazd@deepmind.com}

\author{Leslie Acker}
\affiliation{%
  \institution{DeepMind Technologies, Ltd.}
  \streetaddress{10065 Jasper Avenue, Suite 250}
  \city{Edmonton}
  \state{Alberta}
  \country{Canada}
  \postcode{T5J 3B1}
}
\email{leslieacker@deepmind.com}

\author{Matthew M. Botvinick}
\affiliation{%
  \institution{DeepMind Technologies, Ltd.}
  \city{London}
  \country{United Kingdom}
}
\email{botvinick@deepmind.com}

\author{Joseph Modayil}
\affiliation{%
  \institution{DeepMind Technologies, Ltd.}
  \streetaddress{10065 Jasper Avenue, Suite 250}
  \city{Edmonton}
  \state{Alberta}
  \country{Canada}
  \postcode{T5J 3B1}
}
\email{modayil@deepmind.com}

\author{Adam White}
\affiliation{%
  \institution{DeepMind \& University of Alberta \& Alberta Machine Intelligence Institute}
  \streetaddress{10065 Jasper Avenue, Suite 250}
  \city{Edmonton}
  \state{Alberta}
  \country{Canada}
  \postcode{T5J 3B1}
}
\email{adamwhite@deepmind.com}

\author{Patrick M. Pilarski}
\orcid{0000-0003-1686-2978}
\affiliation{%
  \institution{DeepMind \& University of Alberta \& Alberta Machine Intelligence Institute}
  \streetaddress{10065 Jasper Avenue, Suite 250}
  \city{Edmonton}
  \state{Alberta}
  \country{Canada}
  \postcode{T5J 3B1}
}
\email{ppilarski@deepmind.com}

\begin{abstract}
    Artificial intelligence systems increasingly involve continual learning to enable flexibility in general situations that are not encountered during system training. Human interaction with autonomous systems is broadly studied, but research has hitherto under-explored interactions that occur while the system is actively learning, and can noticeably change its behaviour in minutes. In this pilot study, we investigate how the interaction between a human and a continually learning prediction agent develops as the agent develops competency. Additionally, we compare two different agent architectures to assess how representational choices in agent design affect the human-agent interaction. We develop a virtual reality environment and a time-based prediction task wherein learned predictions from a reinforcement learning (RL) algorithm augment human predictions. We assess how a participant's performance and behaviour in this task differs across agent types, using both quantitative and qualitative analyses. Our findings suggest that human trust of the system may be influenced by early interactions with the agent, and that trust in turn affects strategic behaviour, but limitations of the pilot study rule out any conclusive statement. We identify trust as a key feature of interaction to focus on when considering RL-based technologies, and make several recommendations for modification to this study in preparation for a larger-scale investigation. A video summary of this paper can be found at \href{https://youtu.be/oVYJdnBqTwQ}{https://youtu.be/oVYJdnBqTwQ}.
\end{abstract}

\keywords{Reinforcement Learning, Predictions, Representations, Virtual Reality, Communication}

\AtBeginDocument{%
  \providecommand\BibTeX{{%
    \normalfont B\kern-0.5em{\scshape i\kern-0.25em b}\kern-0.8em\TeX}}}

\begin{teaserfigure}
  \centering
  \includegraphics[width=\textwidth]{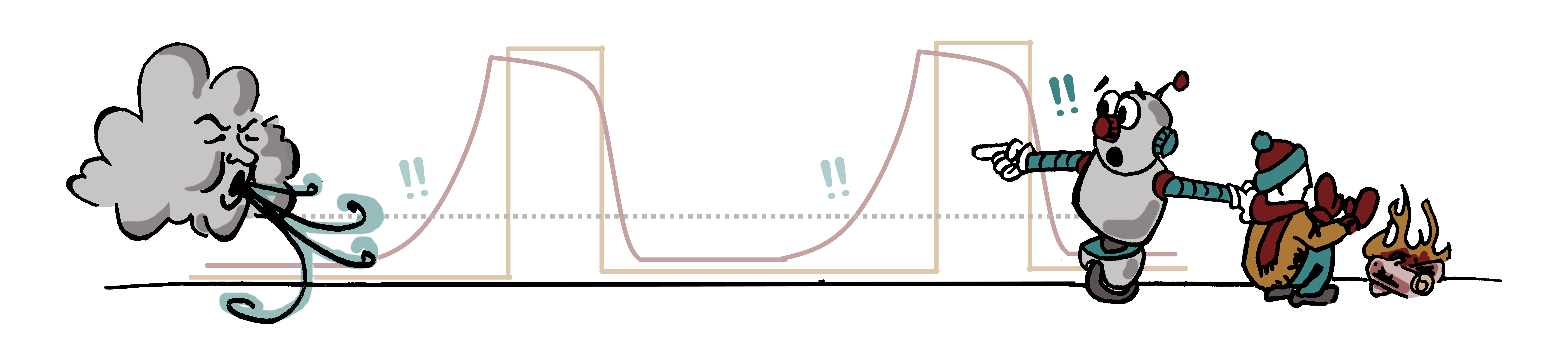}
  \Description{A cartoon representation of a machine agent helping a human avoid a cold wind.}
  \label{fig:teaser}
\end{teaserfigure}

\begin{document}
\pagestyle{fancy}
\fancyhead{}
\maketitle
%%%%%%%%%%%%%%%%%%%%%%%%%%%%%%%%%%%%%%%%%%%%%%%%%%%%%%%%%%%%%%%%%%%%%%
% INTRODUCTION
%%%%%%%%%%%%%%%%%%%%%%%%%%%%%%%%%%%%%%%%%%%%%%%%%%%%%%%%%%%%%%%%%%%%%%
\section{Introduction}
\label{sec:introduction}
 Technology increasingly relies on learning to improve performance. Autonomous systems that continually support human users are expected to soon need to learn continually even during use in order to perform well in their general and changing settings of interest (e.g., assistive technologies;  \cite{adaptive-switching,adaptive-artificial-limbs,dalrymple2020pavlovian,sherstan_thesis}). Humans that use these systems will interact with technology that has a constantly changing level of competency and reliability, but the ramifications of a system's continual learning on human behaviour and human-machine interaction are not well understood. Here, we begin to investigate this interaction by considering a human involved in a timekeeping task, partnered with a machine agent that learns from a blank slate to help the human. In general terms, an intelligent machine of this sort is able to make predictions about the dynamics of the world that a human partner either cannot or does not want to compute on their own (possibly due to the difficulty or time-consuming nature of the computation, \cite{risko2016}, or the human's inability to sense relevant information). In order to convey the benefit of these predictions, it is natural that machine agents must be able to communicate information to the human \cite{lazaridou2020,crandall2018}; learned communication is built upon relationships, and relationships can be built up through interaction \cite{scottphillips2009,scottphillips2014}. Take for example your interactions with a wristwatch: if up to now it conveyed accurate time-information to you, you would have every reason to continue trusting its information the next time you consulted it. If its degree of competency degraded for some reason, and the information communicated were incorrect, you would quickly lose trust in the device and look to other sources for the information you need. Now suppose instead that your wristwatch was not designed to convey regular time intervals, but instead predict the onset of stochastically reoccurring events. How would your interactions with your wristwatch be affected by the fact that the device must continually learn, update, and change its behaviour {\em while you are using it}?

 In this paper we describe a pilot human-agent interaction study, investigating how time-based prediction agents can augment human predictions, and how the relationship between the human and agent develops as the agent develops competency. Specifically, we describe and compare two simple agents that learn to predict future stimuli using {\em general value functions} (\cite{sutton2011}; from the field of reinforcement learning), and communicate those predictions to a human participant using {\em Pavlovian control} (which maps predictions to a small set of actions, \cite{modayil2014prediction}). We introduce a virtual reality (VR) task designed to assess human-agent interaction in a time-interval prediction task. VR is a compelling tool for human-computer interaction (HCI) research because it is immersive, allows flexibility and control for experiment parameters, and enables measurement of human movements which provide a window into decision-making \cite{movement-thinking}. VR also requires physical participation---due to COVID-19, we were unable to recruit external participants. We took this as an opportunity to engage in the present work: a thorough preliminary investigation in search of interesting trends and themes that might deserve careful investigation with respect to continual learning during human-machine interaction.
%%%%%%%%%%%%%%%%%%%%%%%%%%%%%%%%%%%%%%%%%%%%%%%%%%%%%%%%%%%%%%%%%%%%%%
% BACKGROUND
%%%%%%%%%%%%%%%%%%%%%%%%%%%%%%%%%%%%%%%%%%%%%%%%%%%%%%%%%%%%%%%%%%%%%%
\section{Background}
\label{sec:background}
\subsection{Prior Work on Human Interaction with Learning Systems}
\label{sec:prior-work}
Human interaction research regarding autonomous systems spans from early software interfaces for email and calendar applications \cite{Maes1994} to more complex and personal domains such as the control of prosthetic limbs \cite{embodied-cooperation,adaptive-artificial-limbs}, and has included a wide variety of automation techniques. Automation has traditionally been hand engineered to provide reliable performance, and therefore reliable human interaction. More recent machine learning systems are typically pre-trained before deployment, after which their parameters remain fixed. Research specifically involving interaction with continually learning algorithms has hitherto mainly focused on investigating agent learning dynamics using human interaction as part of the learning signal \cite{human-centered-rl-survey}. Autonomous systems that learn from human signals are important technologies, but system learning dynamics are inherently intertwined with interaction dynamics. Amershi et al. \cite{power-to-the-people} convincingly argue the case for separating human interaction from agent learning in order to study ``how people actually interact---and want to interact---with learning systems''. They describe case studies involving people interacting with machine learning systems, and by specifically focusing on the human component of the interaction, they are able to discover novel modes of interaction, unforeseen obstacles, and unspoken assumptions about machine learners. A meta-review of factors that affect trust in human-robot interaction \cite{hri-factors} suggests that system-specific factors such as behaviour, predictability, and failure rates greatly affect human trust in autonomous systems, justifying a system-specific investigation of human interaction with RL-based systems as distinct from machine learning systems. The particular feature of the RL algorithm that we study that distinguishes it from other autonomous systems and warrants direct investigation is continual learning during the course of a task, and the effect that will have on human interaction.
%%%%%%%%%%%%%%%%%%%%%%%%
%% SUBSEC: GVFS
%%%%%%%%%%%%%%%%%%%%%%%%
\subsection{General Value Functions}
\label{sec:gvfs}
Reinforcement learning \cite{sutton2018reinforcement} is a class of machine learning methods wherein an agent learns to predict future values through a process of trial-and-error. The value of a state (a prediction of how much reward can be expected in the future from that state) is learned by incremental updates to a value function $v_\gamma(s)$ for state $s$. The discounting factor $\gamma$ corresponds to the horizon of the prediction, and is typically between 0 (for next-step predictions) and 1 (for an infinite horizon). By substituting any signal of interest (called a {\em cumulant}, $C$) in place of the reward, the value function becomes a {\em general value function} (GVF) $v_{\gamma,C}(s)$ which predicts the discounted sum of the future cumulant \cite{sutton2011,nexting}.  Informally, a GVF represents a prediction question: {\em what will be the total accumulated value of some signal of interest over the next specified time window?} Equation \ref{eq:gvf} gives the formal GVF definition, for a simple fixed-$\gamma$ on-policy prediction formulation. 
\begin{equation}
\label{eq:gvf}
v_{\gamma,C}(s) = \mathbb{E} \Bigg \{ \sum_{k=0}^\infty \gamma^k C_{t+k+1} \; \bigg | \; S_t = s \Bigg \}
\end{equation}
In practice, an agent learns to approximate the
above value by interacting with a stream of states and corresponding cumulants. Let $x(s) \in \mathbb{R}^d$ be a feature vector summarizing the state $s$.  We  approximate the value by $v_{\gamma,C}(s) \approx w_t ^\intercal x(s)$, where $w_t \in \mathbb{R}^d$ is the weight vector at time $t$. We use the TD($\lambda$) algorithm to update $w_t$ on each time step: 
\begin{align*}
e_t &= e_{t-1} + x(S_{t})\\
\delta_t &= C_{t+1} + \gamma w_t ^\intercal x(S_{t+1}) -  w_t ^\intercal x(S_{t})\\
w_{t+1} &= w_t + \alpha \delta_t e_t\\ 
e_{t} &= \gamma\lambda e_{t},
\end{align*}
where $\alpha$ is a scalar learning rate and $e \in \mathbb{R}^d$ is an exponentially decaying memory of previous feature activations.
%%%%%%%%%%%%%%%%%%%%%%
%% SUBSEC: PAVLOVIAN CONTROL
%%%%%%%%%%%%%%%%%%%%%%
\subsection{Pavlovian Control}
\label{sec:pavlovian}
Inspired by prediction learning for reflexive control in animals \cite{Kehoe2002}, the term {\em Pavlovian control} as used here refers to the use of a GVF to predict an external stimulus, coupled with a fixed reflexive control policy dependent on that prediction (c.f., \cite{modayil2014prediction,dalrymple2020pavlovian}). A simple Pavlovian control policy emits a discrete action $a_1$ when the GVF prediction of the external stimulus is below a certain threshold $\tau$, and emits a discrete action $a_2$ otherwise (where, importantly, that action may be a communication signal \cite{parker2019,pilarski2021}). The precise setting of $\tau$ for a useful policy depends on the timescale and stochasticity of the prediction, as well as the amount of advance notice needed before the external stimulus in order to take action.
%%%%%%%%%%%%%%%%%%%%%%%%%%%%%%%%%%%%%%%%%%%%%%%%%%%%%%%%%%%%%%%%%%%%%%
% METHODS
%%%%%%%%%%%%%%%%%%%%%%%%%%%%%%%%%%%%%%%%%%%%%%%%%%%%%%%%%%%%%%%%%%%%%%
\section{Experimental Methods}
\label{sec:methods}
Our experiment situates a human participant in a virtual reality (VR) environment we call {\em the Frost Hollow}, wherein they must keep track of an external event that occurs on a roughly periodic schedule (c.f., \cite{rafiee2021}). They are paired with a machine agent that uses a GVF to predict the onset of this event, and cues the human when its prediction exceeds a threshold. We look at task performance, behavioural differences, and qualitative notes to compare two agent architectures against the control condition where the participant completes the task with no agent assistance.
\subsection{Virtual Reality Environment}
\label{sec:vr}
\begin{figure}
    \centering
    \subcaptionbox{Participant view\label{fig:vr-first-person}}
    {\includegraphics[height=1.6in]{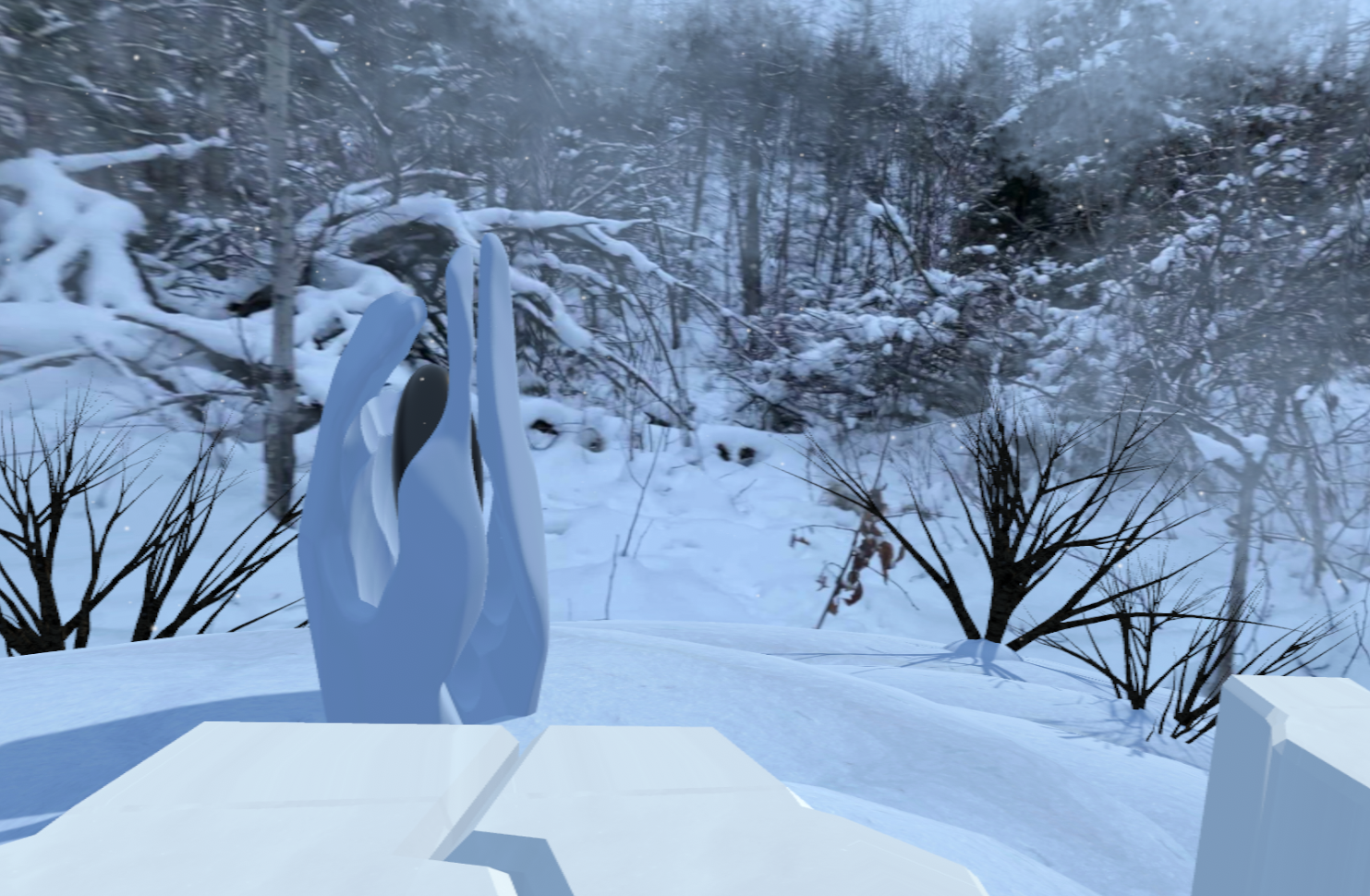}} 
    \subcaptionbox{Top-down view, annotated\label{fig:vr-env}}
    {\includegraphics[height=1.6in]{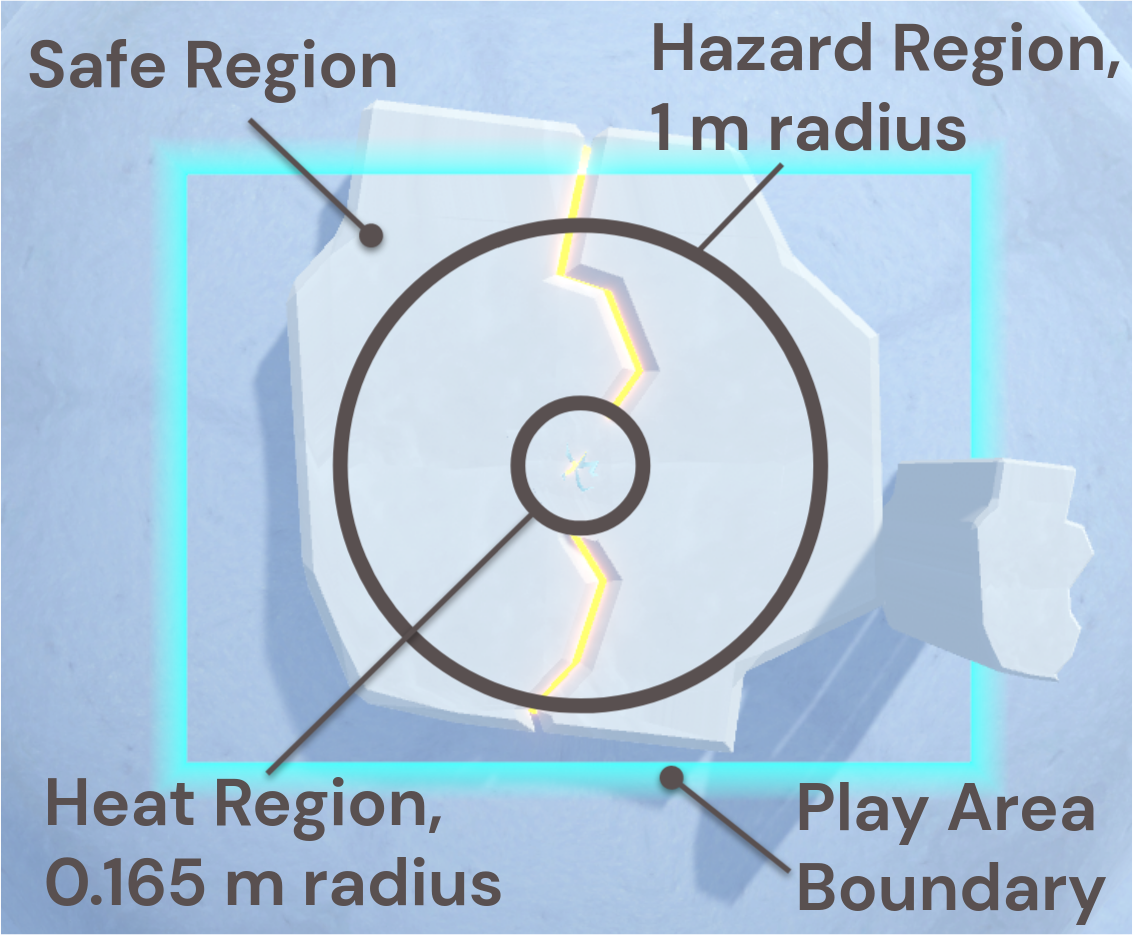}}
    \caption{Depiction of the virtual reality environment.}
    \label{fig:vr}
    \Description{A virtual-reality environment made to emulate a beautifully snowy frost hollow. Snow covers the ground and surrounding trees. A ice-sculpture heat-gauge is visible left of center. In the right figure, a top-town view is presented, annotating a center circle with radius 0.165 m corresponding to the heat generation region, and a larger 1 m radius circle corresponding to the hazard region.}
\end{figure}

The premise of the Frost Hollow task is that the participant stands in a ``warm'' center region of the environment (radius 0.165 m, participant position reported by the headset) to slowly gain heat, and must periodically dodge out of a hazard region (radius 1 m) when the wind blows to avoid losing heat. When standing in the center region, a heat gauge visible to the participant fills from 0.0 to 5.0 at a rate of 0.1875 heat/second (26.67 seconds to fill the gauge); when the gauge is full, the participant can raise one of their VR controllers above their headset to cache the heat gained as a point (one unit of game reward). When hit by the hazard, the participant loses 25 heat/second, so any hit longer than 200 ms will deplete the gauge. Cached points are not lost. Our VR environment (depicted in Figure \ref{fig:vr}) was implemented in Unity 2019.2.17f1 (Unity Technologies, USA) with Steam VR (Valve Corporation, USA) and presented to the participant via a Valve Index headset and two handheld controllers (Valve Corporation, USA; headset max render rate of 144 Hz) at a base Unity time step length of approximately 8ms (VR protocol follows from prior work \cite{pilarski2019}).  Detailed descriptions of the audiovisual presentation of the environment are available from Pilarski et al. \cite{pilarski2021}. We studied three inter-stimulus-interval (ISI) conditions between the hazard pulses: fixed, drifting, and random. The base ISI was set to 20s (as measured from the falling edge of the pulse to the next pulse's falling edge); the hazard pulse duration was 4s in all conditions. For the random condition, the inactive portion of each ISI was varied uniformly by [-4s, 6s] between 12s and 22s in length. For the drift condition, the inactive portion of each ISI was shifted by a uniform random amount between [-2s, 2s] from the previous duration, with all ISI durations outside [12s, 22s] cropped to the extremes of the range. When the hazard pulse was active, the participant's left hand-held controller vibrated, and a visual bloom stimulus was presented on hazard contact; communication from the agent was presented as vibration in the right hand-held controller (c.f., \cite{parker2019,edwards2016}).
\subsection{Agent Architectures}
\begin{figure}[b]
    \centering
    \subcaptionbox{Bit Cascade Representation\label{fig:bc-representation}}
    {\includegraphics[width=0.4\textwidth]{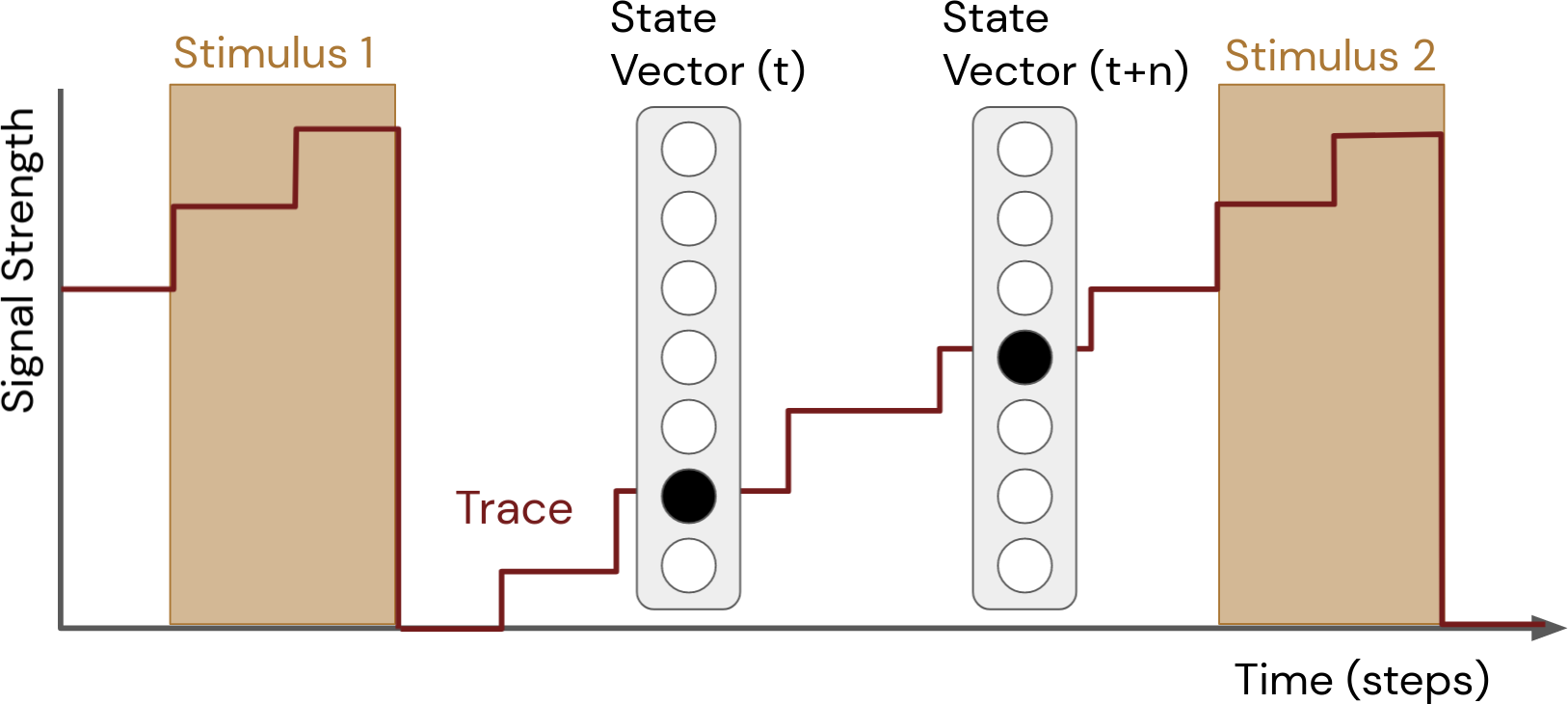}}
    \subcaptionbox{Tile-Coded Trace Representation\label{fig:tct-representation}}
    {\includegraphics[width=0.4\textwidth]{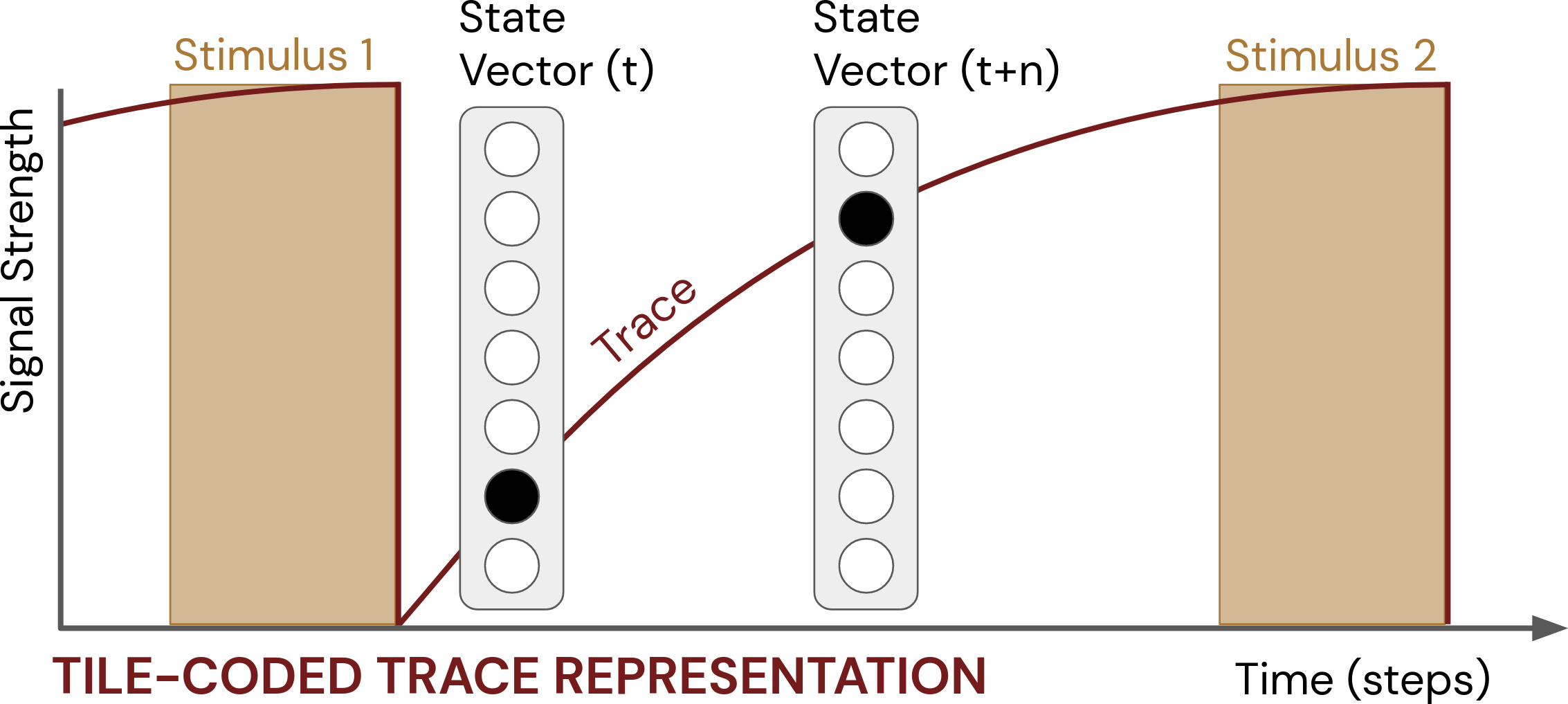}}
    \caption{Representations of time used in this experiment. Time (state) is represented as a one-hot vector of features which activate according to a trace function which resets at the falling edge of each stimulus pulse.}
    \label{fig:representations}
    \Description{Two charts depicting the agents' representations of time. The bit cascade representation shows a staircase-like function rising from one stimulus to the next, while the tile-coded trace representation shows a smooth exponential curve rising from one stimulus to the next. Both representations show a one-hot vector with active features corresponding to the value of the trace at a given time.}
\end{figure}
Two agent architectures are compared which differ only in the way that they represent the passage of time between stimuli: a {\em bit cascade} (BC) representation, and a {\em tile-coded trace} (TCT) representation (depicted visually in Figure \ref{fig:representations}). The decision to vary agent representations of time rather than other agent parameters is motivated by a larger study of Pavlovian signalling \cite{pilarski2021} for which this work plays a supporting role. These representations of time were motivated by and modeled after biological models of time-keeping in animal brains \cite{paton2018}. The BC representation is modelled after population clocks (sequentially firing chains of cells), while the TCT representation is informed by ramping models (changes in the tonic firing rate of cells or cell populations). The bit cascade representation involves a one-hot vector of 40 features which activate sequentially, with each feature being active for 0.5s. The tile-coded trace representation also involves a one-hot vector of 40 features which activate sequentially, but the activation timing for each feature is prescribed by an exponential decay trace with a per-step decay rate of 0.998. Both the BC and TCT representations restart their sequence (i.e. the first feature is active) immediately after the hazard pulse deactivates. The timing parameters for both representations were set so that both used roughly the same number of feature bins when presented with an ISI of 20s. Learning parameters were empirically determined for an acceptable learning speed over a 5 minute trial time, resulting in $\alpha = 0.1$, $\lambda = 0.99$, and $\gamma = 0.99$. The Pavlovian control threshold $\tau$ was also empirically determined to give adequate lead-time for a human participant in advance of a pulse after learning had converged, resulting in $\tau = 10$ for both agents. The fixed control policy was set such that the agent vibrated the participant's handheld controller when its prediction rose above $\tau$, and did not vibrate when below $\tau$. Agent-learned weights were discarded and re-initialized to zero between trials so the agent learned from a blank slate for each trial.
\subsection{Experiment and Analysis Protocol}
\label{sec:protocol}
We engaged a single participant (male, age 40, no history of sensorimotor impairments) {\em who was also a member of the study team} due to COVID-19 constraints (see Section \ref{sec:limitations}), and followed our approved human research ethics protocol. This participant had a deep understanding of the task and dynamics, but was not practiced with the particular conditions. The study followed a within-participant 3 (ISI type) x 3 (agent type) design; experimentation took place over the course of ten sessions, each consisting of nine trials that were five minutes long (one for each pairwise combination of [fixed ISI, drifting ISI, random ISI] and [no agent, TCT agent, BC agent]). Trial order was randomized and blinded to the participant, and the initial ISI duration for the fixed and drifting conditions was randomized to further obfuscate the conditions. Each individual session was conducted in roughly one hour, with small breaks between each of the trials for the participant to remove the headset and drink water or write qualitative notes. Sessions were spread over a one month collection period, with one or two sessions per day on data collection days. This protocol was found to be slightly physically fatiguing and moderately cognitively fatiguing, depending on the trial. To avoid injecting biases into the analysis, the team member who acted as participant for the study did not re-engage with the study until both qualitative and quantitative analyses were completed by other team members. Statistical analyses were conducted to determine whether for this participant there were any differences in performance across agent types. Data violated assumptions of normality in nearly every comparison, so non-parametric methods were used. Data were grouped pair-wise by session, so Friedman's tests were conducted followed by Wilcoxon Signed-Rank tests with a Holm-\v{S}id\'{a}k correction for multiple comparisons. Significance is reported in Figure \ref{fig:performance-metrics}, in all cases at the family-wise $\alpha=0.05$ level. Specific results of the statistical analyses are reported in Table \ref{tab:stats}.
%%%%%%%%%%%%%%%%%%%%%%%%%%%%%%%%%%%%%%%%%%%%%%%%%%%%%%%%%%%%%%%%%%%%%%
% QUANTITATIVE ANALYSIS
%%%%%%%%%%%%%%%%%%%%%%%%%%%%%%%%%%%%%%%%%%%%%%%%%%%%%%%%%%%%%%%%%%%%%%
\section{Quantitative Analysis}
\label{sec:quant}
\begin{figure}[b]
    \centering
    \subcaptionbox{Overall task performance\label{fig:pointscached}}
    {\includegraphics[width=0.33\textwidth]{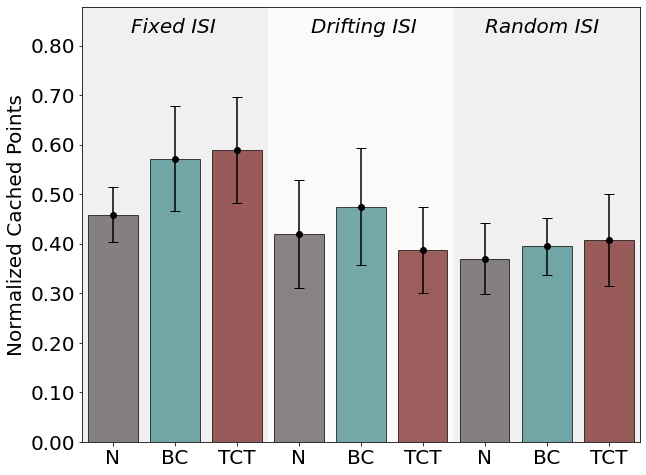}}
    \subcaptionbox{Time-Steps hit by the hazard\label{fig:hitsteps}}
    {\includegraphics[width=0.33\textwidth]{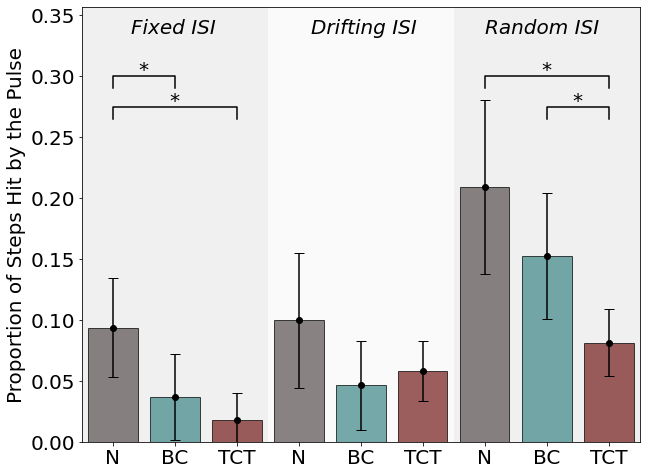}}
    \subcaptionbox{Total heat gain\label{fig:heatgain}}
    {\includegraphics[width=0.33\textwidth]{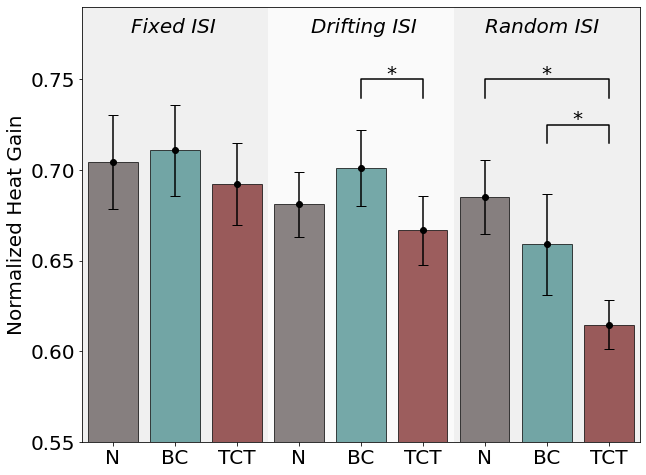}}
    \caption{Performance metrics. Bars represent the mean over trials for each metric, normalized by the maximum possible value of that metric. Error bars represent the 95\% confidence interval. N = no agent; BC = bit-cascade agent; TCT = tile-coded trace agent.}
    \label{fig:performance-metrics}
    \Description{Bar charts depicting performance metric differences across conditions. There are no significant differences in overall task performance. When paired with either agent, the participant is hit by the hazard less than with no agent in the fixed ISI condition. In the random ISI condition, the participant is hit by the hazard less with the tile-coded trace agent than with the bit-cascade agent, or no agent. The participant gains less heat with the tile-coded trace agent than with the bit-cascade agent in either the drifting ISI or random ISI conditions. The participant also gains less heat with the tile-coded trace agent than with no agent in the random-ISI condition.}
\end{figure}
\begin{table*}[t]
    \caption{Statistical analysis results. Comparisons are made across assistant pairings (N = no agent; BC = bit-cascade agent; TCT = tile-coded trace agent) for each ISI condition. Significance ($\alpha=0.05$) is indicated in bold text. For Friedman's tests, $\chi^2_{critical}(2)=6.20$.}
    \label{tab:stats}
    \begin{tabular}{lrccc}
    \toprule
    & & Points Cached & Steps Hit by Hazard & Heat Gain \\
    \midrule
    \multicolumn{5}{l}{\bf{A Priori Tests (Friedman's Chi-Square)}}\\
    Fixed & $\chi^2(2)$ & 5.2 & \bf{14.0} & 1.4  \\
    & $p$ & 0.0755 & \bf{0.0009} & 0.4966  \\ 
    Drifting & $\chi^2(2)$ & 1.4 & 1.4 & \bf{7.2}  \\
    & $p$ & 0.4895 & 0.4966 & \bf{0.0273}  \\ 
    Random & $\chi^2(2)$ & 0.7 & \bf{11.4} & \bf{13.4}  \\
    & $p$ & 0.7165 & \bf{0.0033} & \bf{0.0012}  \\ 
    \multicolumn{5}{l}{\bf{Post Hoc Tests (p-values from Wilcoxon Signed-Rank)}}\\
    Fixed & N vs BC & 0.1415 & \bf{0.0432} & 0.9594 \\
    & N vs TCT & 0.1724 & \bf{0.0151} & 0.5550  \\
    & BC vs TCT & 0.7344 & 0.1763 & 0.4930 \\
    Drifting & N vs BC & 0.6831 & 0.4257 & 0.3642 \\
    & N vs TCT & 0.6831 & 0.4881 & 0.3642  \\
    & BC vs TCT & 0.5507 & 0.6465 & \bf{0.0206}  \\
    Random & N vs BC & 0.7990 & 0.2026 & 0.0593  \\
    & N vs TCT & 0.7990 & \bf{0.0278} & \bf{0.0151}  \\
    & BC vs TCT & 0.7990 & \bf{0.0278} & \bf{0.0329}  \\
\bottomrule
\end{tabular}
\end{table*}
Looking first at overall task performance (Figure \ref{fig:pointscached}), we see a small (and not statistically significant) increase in performance in the fixed ISI condition when the participant was paired with either agent. For the more difficult conditions where the ISI changes over the course of the trial, there is no clear difference in overall task performance depending on agent pairing. In general, these results suggest that overall task performance is not a clear indicator of any differences between human-agent pairings in this setting. Figure \ref{fig:hitsteps} shows differences in the proportion of time-steps where the participant was hit by the hazard. In the fixed ISI condition, the participant spends less time being hit by the pulse when paired with either agent as compared to none. In the random ISI condition, the participant is hit by the pulse less when paired with the TCT agent than when paired with the BC agent, or no agent. Figure \ref{fig:heatgain} displays the participant's heat gain in each condition, which corresponds to the proportion of time spent in the goal region. Differences here appear in the more challenging conditions, where the participant spends less time in the goal region when paired with the TCT agent than when paired with the BC agent. Considering the charts of Figure \ref{fig:performance-metrics} together, it appears that the participant engages in more cautious behaviour when paired with the TCT agent as compared to the BC agent (they gain less heat, and are hit by the hazard less often), while attaining comparable task performance. This result suggests possible differences in participant behaviour across agent pairings. 
\begin{figure*}
    \centering
    % SUBFIG: SIGNAL TO PULSE INTERVALS
    \subcaptionbox{Mean interval between signal from agent and hazard onset over trial length. The minimum useful lead time (dashed line) before the hazard pulse (dotted line) is 0.89 seconds, and corresponds to the participant's mean exit speed. It does not include reaction time.\label{fig:sig-timing}}
    {\includegraphics[width=\textwidth]{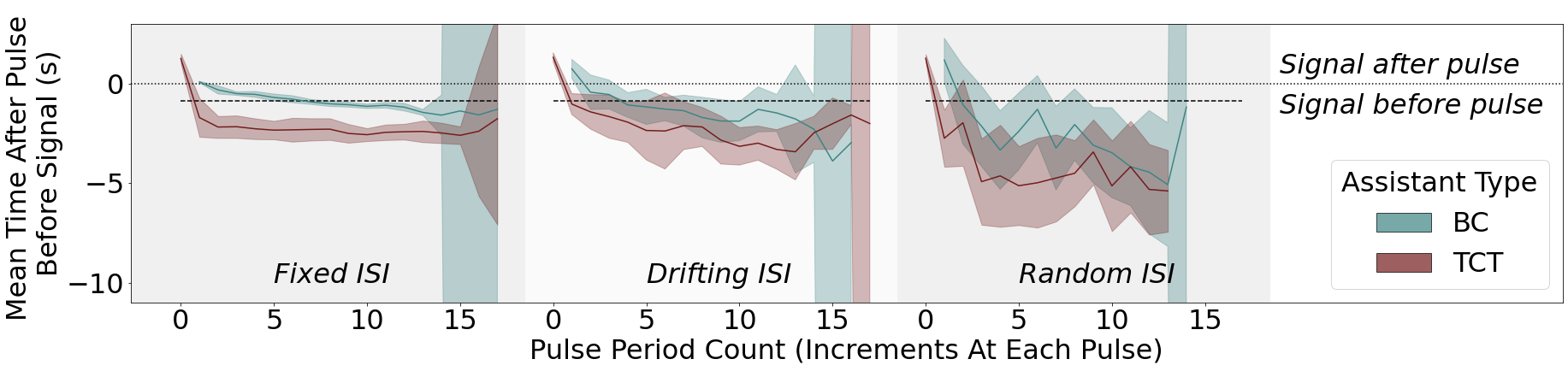}}
    \Description{Line charts depicting the mean interval between the agent's signal and the onset of the hazard pulse. In all conditions, the tile-coded trace agent provides signals sooner and more reliably than the bit-cascade agent. More difficult conditions (drifting and random ISIs) exhibit more variability, but the trend remains.}
    % SUBFIG: SIGNAL TO EXIT INTERVALS
    \subcaptionbox{Time interval between signal from agent and goal region exit, shown as a trajectory over the length of trials. Negative data indicate the participant leaving the goal region in advance of the agent's signal.\label{fig:sig-to-exit-intervals}}
    {\includegraphics[width=\textwidth]{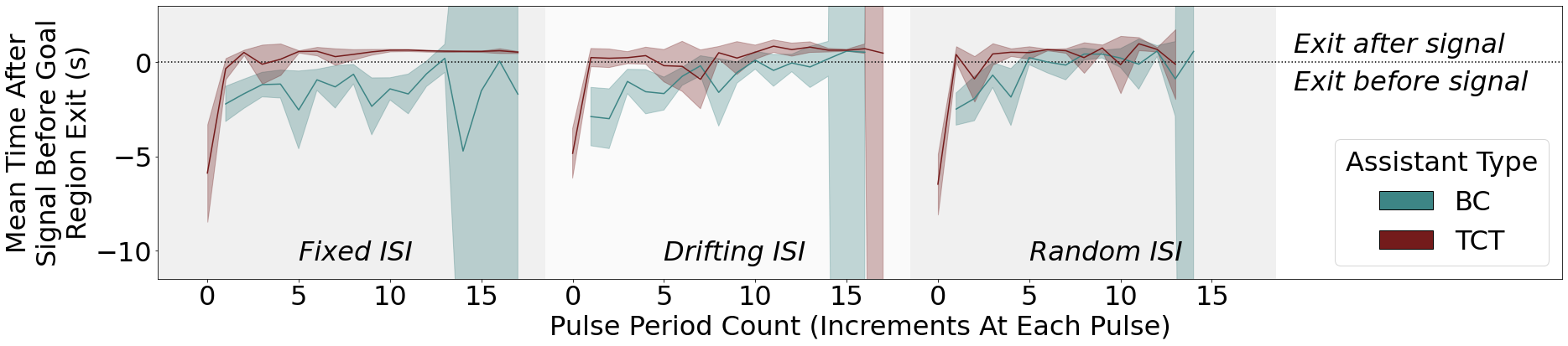}}
    \Description{Line charts depicting the mean interval between the agent signal and the participant's exit from the goal region. In all conditions, the participant aligns their exit to follow the agent signal sooner in the trial and more reliably with the tile-coded trace agent than with the bit-cascade agent.}
    % OVERALL FIGURE
    \caption{Data are shown as the mean (solid line) and 95\% confidence interval (shaded region) of the data for each pulse. Due to the randomization of the starting ISI and fixed trial duration, some trials with shorter ISIs presented more pulses than others. This led to the occurrence of one or two trials with high pulse count (>14), resulting in the large confidence intervals at the ends of these plots.}
\end{figure*}
 
 We are able to examine agent learning directly because the agent's learning of the task does not depend on the participant's actions. Figure \ref{fig:sig-timing} shows the mean interval between the agent signal and the onset of the hazard pulse, for each pulse over the length of the trials. This interval can be interpreted as ``how long before the onset of the pulse did the agent's prediction of the pulse rise above the threshold for signalling?''. When the agent's cue is less than 0.89s before the hazard (above the dashed line), the signal doesn't give the participant enough time to react given how long it takes to leave the hazard region. In the simplest prediction task with fixed ISI, the BC agent is unable to reliably give a useful signal (below the dashed line) until after about the sixth or seventh pulse while the TCT agent is able to give a reliably useful signal after only the second pulse. More challenging conditions introduce more variance in these intervals, but the trend remains that the TCT agent provides useful signals earlier, and more reliably. This is because the BC representation has finer feature bins in the region near the pulse compared to the TCT representation, leading to more accurate but slower learning.
 
 The length of time between the agent's signal and the participant's exit from the goal region is plotted in Figure \ref{fig:sig-to-exit-intervals}. A negative value on this chart indicates that the participant left the goal region before being cued by the agent. Here, we see that the participant exhibits clear behavioural differences when interacting with each agent. When paired with an agent with a TCT representation, the participant nearly always waits for the agent signal before leaving the goal region (data above the dotted line). When paired with an agent with a BC representation, the participant is much more likely to exit the goal region before the agent gives a signal. In the fixed ISI condition, when paired with the TCT agent, the participant seems to move after the agent's cue as early as the second or third pulse of a trial. Under the same conditions, when paired with a BC agent, the participant relies entirely on their own internal timing. For the more difficult conditions, the participant eventually moves after the cue of either agent, but aligns their movements with the TCT agent's cue more readily than with the BC agent's cue. While it is tempting to interpret this feature of the data as the participant relying on the TCT agent's cue more than the BC agent's cue, there is insufficient evidence from these charts alone to conclude how the participant is using either signal, as we will see in Section \ref{sec:qual}.
%%%%%%%%%%%%%%%%%%%%%%%%%%%%%%%%%%%%%%%%%%%%%%%%%%%%%%%%%%%%%%%%%%%%%%
% QUALITATIVE ANALYSIS
%%%%%%%%%%%%%%%%%%%%%%%%%%%%%%%%%%%%%%%%%%%%%%%%%%%%%%%%%%%%%%%%%%%%%%
\section{Qualitative Analysis}
\label{sec:qual}
Qualitative data was gathered by the participant after each session in free-form text, prompted but not restricted by the following questions. Experimenter-developed questions were posed by the member of the study team conducting the qualitative analysis at the outset of the trials. Participant-developed questions were generated independently by the member of the study team acting as the participant, and evolved as the study progressed.\\
{\em Experimenter-developed questions:}
\begin{itemize}
    \item Are you trying to figure out how the agent (and environment) work?
    \item For the whole trial?
    \item If not, did you figure it out or just start to trust it?
    \item After time, or successes? 
    \item How much do you notice or think about the other agent at the beginning? The middle? The end?
\end{itemize}
{\em Participant-developed questions:}
\begin{itemize}
    \item Changes in when and how I counted: did I count from the start of the trial? Did I shift to just counting from the agent cue and not counting from the beginning? When did I shift between these and under what conditions or observations on timing?
    \item What agent behaviours did I like and not like?
    \item Adaptation rates: what were my expectations on response or learning times for agents?
    \item Thinking of agents as adaptive systems / predictors or not?
    \item What conditions did I lose confidence in the machine; when did I gain confidence?
    \item When did trust in the agent occur quickly?
\end{itemize}
Discourse analysis seeking recurring sentiment and themes indicated that trust was a major component of the participant's interaction with the system, which affected other factors including cognitive load and use of the agent's cue in unexpected ways. The participant noted that they trusted the agent more when it was demonstrably correct earlier in the trial. Once trust was built, they noticed a decrease in cognitive load: ``{\em With trust in my agent, I can let [my] mind wander}''. Notes such as ``{\em [The] agent helps me feel like I have a lower bound of safety once it is trained, and then can choose my risk based on its feedback}'' suggest that the participant engaged with the task actively and strategically, and used the agent's cues as part of that strategy in more complex ways than rote cue-to-movement. In fact, with sufficient experience with the agent, the participant would sometimes engage in risky behaviour: ``{\em I was at times racing the pulse; the agent would cue me but I would see the heat bar almost full and then gamble that it would fill fast enough before [the hazard] came, given what I knew about the relationship between cue and future pulse.}'' Even in cases where the agent inadequately predicted the hazard, the participant still used the agent signal as information to inform their strategy, but relied on their mental timekeeping to inform their movements. Regarding these situations the participant notes: ``{\em it was not fast enough to be useful in advance of [the hazard], so I mainly used it as a checksum}'', indicating that they verified their mental timekeeping by comparing it against their acquired knowledge of how the agent keeps time. While this particular behaviour is likely unique to participants familiar with GVF agents of this nature, the anecdote provides a clear example of how a user's mental model of an agent will affect their interactions with it. It also indicates that evaluating future participants' understanding of how the agent learns will be key to understanding their interactions.
%%%%%%%%%%%%%%%%%%%%%%%%%%%%%%%%%%%%%%%%%%%%%%%%%%%%%%%%%%%%%%%%%%%%%%
% DISCUSSION
%%%%%%%%%%%%%%%%%%%%%%%%%%%%%%%%%%%%%%%%%%%%%%%%%%%%%%%%%%%%%%%%%%%%%%
\section{COMPARING QUANTITATIVE AND QUALITATIVE RESULTS}
\label{sec:qual-and-quant}
 In both the quantitative and qualitative analyses we see human trust of the agent emerging as an important theme. The participant's notes suggest that using the sign of the signal-to-exit interval (Figure \ref{fig:sig-to-exit-intervals}) as an indicator of human trust might miss parts of the picture, since the participant makes use of the agent signal in other ways than as simply a cue to move. Other quantitative measures of trust should be sought, to corroborate this interpretation. One particular notion of intense trust called out in the participant notes (when the participant is ``racing'' the pulse, caching points after the agent signal but before the hazard) is also visible in the quantitative data. Of the 14 instances where a point caching event is recorded after an agent signal and before a hazard, 13 of these instances occurred when the participant was paired with the TCT agent. This risky behaviour with the TCT agent contrasts with the indications from Figure \ref{fig:performance-metrics} that the participant behaved more cautiously with the TCT agent. Pairing this contrast with the qualitative discussion, we see that with high levels of trust in the agent, the participant is able to more flexibly choose a strategy, behaving boldly or cautiously as the situation warrants. It should however be noted that (as shown in Figure \ref{fig:sig-timing}) the TCT agent reliably gives more lead-time than necessary before the pulse, leaving time for pulse racing that the BC agent does not, meaning that pulse racing may not be a fair indicator of trust. 
 
 \section{Discussion}
 \label{sec:discussion}
 Specific quantitative and qualitative measures to assess human trust in the agent would be particularly informative for future studies, especially if such measures could assess changes in levels of trust over the course of a trial or across sessions. One such task modification might involve the introduction of a secondary, voluntary and cognitively demanding task that could be performed simultaneously while gathering heat. While engaged with the secondary task, the participant would need to place trust in the agent to keep track of the timing in the primary task (i.e., effect a form of cognitive offloading \cite{risko2016}), making engagement in the secondary task a good measure of trust.

 Supposing that future studies with a direct measure of trust confirm that participants trust the TCT agent more than the BC agent, two points of discussion emerge. First, the apparent differences in levels of trust between the types of agent can only be attributed to the different representations of the agents, as all else is equal. While the BC agent is able to achieve greater prediction accuracy than the TCT agent (because of the BC agent's finer feature bins in the region of the hazard), fast learning appears to be more important than accuracy for the development of human trust. The threshold and representation bin-widths in this experiment were chosen considering late-trial performance, so that once the predictions stabilized both representations would give roughly equal notice before a pulse. A lower threshold or wider feature bins would likely have allowed the BC agent to provide reliably useful signals earlier in the trials. Understanding the relationship between feature representation and threshold levels in both early and late-learning contexts will be important for any future studies or applications making use of Pavlovian signalling for communication. Second, the participant in this experiment displayed more richly varied strategies with the TCT agent than the BC agent, presumably because of a greater degree of trust. Specific assessments regarding how participant strategies are affected by trust in the agent may be illuminating, and should involve specific metrics to assess changing strategies over time. Finally, using a pre-trained agent as a baseline comparison will be necessary to assess the effect of active learning on these measurements.
%%%%%%%%%%%%%%%%%%%%%%%%%%%%%%%%%%%%%%%%%%%%%%%%%%%%%%%%%%%%%%%%%%%%%%
% LIMITATIONS
%%%%%%%%%%%%%%%%%%%%%%%%%%%%%%%%%%%%%%%%%%%%%%%%%%%%%%%%%%%%%%%%%%%%%%
\section{Limitations}
\label{sec:limitations}
 The generality of this pilot study is limited by our use of a single participant who was also a member of our study team. While blinded from the particular conditions they were interacting with, they were deeply familiar with the agent architectures, task dynamics, and learning machines in general. We expect that the introduction of na\"{i}ve participants will also involve a co-learning phase at the beginning of sessions where the participant and agent are both learning the task simultaneously. Since we found early interactions to be of great influence in trust-building with our expert participant, we expect that a co-learning phase will affect trust, but make no hypothesis about what that effect might be.
%%%%%%%%%%%%%%%%%%%%%%%%%%%%%%%%%%%%%%%%%%%%%%%%%%%%%%%%%%%%%%%%%%%%%%
% CONCLUSIONS/FUTURE WORK
%%%%%%%%%%%%%%%%%%%%%%%%%%%%%%%%%%%%%%%%%%%%%%%%%%%%%%%%%%%%%%%%%%%%%%
\section{Conclusions and Future Work}
\label{sec:conclusions}
 This pilot study examined an approach to agent-human support characterized by real-time machine learning and straightforward ongoing interactions; our results suggest that trust in the system's capabilities is a major component of a human's interaction with a continually learning system. There are also indications that this trust may be dependent mainly on early interactions with the system, while the agent is still developing competency. Future studies should include metrics that specifically measure trust, and should include analyses to determine possible correlations between levels of trust and agent competence. There may also be correlations between levels of trust and strategic behaviour. Finally, a future study should include a greater number of participants, with a diversity of experience in interacting with learning machines. \balance For other future time-based prediction experiments or applications involving human actors with machine agents, we make no particular recommendations about representation or threshold choices, as we understand these to be task-specific. We do however stress the importance of these choices, and recommend that they be made with both early and late learning stages in mind, and considering the interaction between the human and machine's actions.

\begin{acks}
The authors thank Kevin McKee, Martin Riedmiller, Nathan Wispinski, and Ola Kalinowska for many helpful conversations, discussions, and technical insights regarding this work. University of Alberta collaboration on this work by PMP, ASRP, and AW was supported, in part, by the Canada CIFAR AI Chairs program, the Canada Research Chairs program, the Natural Sciences and Engineering Research Council (NSERC), and the Alberta Machine Intelligence Institute (Amii).
\end{acks}
\bibliographystyle{ACM-Reference-Format}
\bibliography{main}

\end{document}